\documentclass[10pt,onecolumn]{article}

\usepackage[margin=0.75in]{geometry}
\usepackage{amsmath,amssymb,amsfonts}
\usepackage{graphicx}
\usepackage{booktabs}
\usepackage{cite}
\usepackage{hyperref}
\usepackage{caption}
\usepackage{textcomp}
\usepackage{cite}
\usepackage{algorithm}
\usepackage{algpseudocode}
\usepackage{multicol}
\usepackage{multirow}

\title{Combinatorial Optimization of Robotic Hand Kinematic Structures Using a Potential-Dexterity-Based QUBO Formulation}

\author{
    HyoJae Kang$^{1}$, Yeong Jae Park$^{1}$, Jeongdo Ahn$^{1}$, and Dong Il Park$^{1,2,*}$ \\
    $^{1}$Advanced Robotics Research Center, Korea Institute of Machinery \& Materials (KIMM) \\
    $^{2}$Mechanical Engineering, University of Science \& Technology (UST) \\
    $^{*}$Corresponding author
}

\begin{document}
\date{}
\maketitle

% ======================
% Footnote (중요)
% ======================
\thispagestyle{plain}
\footnotetext{This manuscript has been submitted for possible publication.\\
This research was supported by the National Research Council of Science \& Technology(NST) granted by the Korea Government(NSIT) (No. GTL25041-000) and by 'Creation of the quantum information science R\&D ecosystem (based on human resources)' through the National Research Foundation of Korea (NRF) funded by the Korean government (RS-2023-NR057243).}

% ======================
\begin{abstract}
Robotic hand kinematic design involves combinatorial selection among multiple finger structures whose performance depends on both individual kinematic characteristics and inter-finger interactions. This study presents a potential-dexterity-based quadratic unconstrained binary optimization (QUBO) formulation for combinatorial optimization of robotic hand kinematic structures. Potential dexterity-related metrics were evaluated through kinematic analysis and encoded into a 27-variable QUBO problem incorporating individual finger performance, thumb-to-finger workspace-overlap interactions, one-hot selection constraints, and pairwise structural compatibility. Exhaustive evaluation of the design space established a reference optimum and enabled direct validation of the formulated objective. Discretization sensitivity analysis showed that the selected design remained unchanged across the tested joint-space sampling intervals and at a finer workspace voxel resolution, while sensitivity was observed under coarse voxelization. Simulated annealing consistently recovered the reference optimum at higher sampling counts, and direct execution on D-Wave quantum annealing hardware also recovered the reference optimum under all tested sampling conditions. These results demonstrate that the proposed formulation provides a common QUBO representation for constrained robotic hand kinematic structure selection and enables evaluation using exhaustive, classical annealing, and quantum annealing approaches.
\end{abstract}

% ======================
\section{Introduction}
The kinematic structure of a robotic hand determines fundamental characteristics such as reachable workspace, manipulability, degrees of freedom (DoF), and potential inter-finger reachability. When several structural parameters are considered simultaneously, including finger link proportions, joint positions, DoF configurations, and palm joints, the number of possible hand configurations increases rapidly. Robotic-hand kinematic design can therefore be regarded as a combinatorial design problem in which multiple structural candidates must be selected while considering both the performance of individual fingers and interactions among them.

Previous studies have evaluated and optimized robotic kinematic structures using workspace and other kinematic performance measures. These include workspace optimization based on position-level kinematic analysis \cite{r1}, optimization considering kinematic sensitivity and workspace performance \cite{r2}, and robotic-hand design approaches considering finger interactivity \cite{r3}, thumb opposability \cite{r4}, and thumb configuration \cite{k4}. Individual performance measures and interaction-based metrics have also been considered within an integrated design evaluation framework \cite{p2}. Such studies demonstrate that kinematic analysis can provide quantitative criteria for comparing candidate structures before considering a specific manipulation task. However, robotic-hand design involves not only evaluating individual structures but also selecting compatible combinations of multiple finger designs while accounting for interactions between them.

This simultaneous selection becomes increasingly difficult as additional discrete design variables are introduced. Combinatorial optimization problems are generally challenging to formulate and solve as the number of possible combinations increases \cite{r5}. Metaheuristic methods can be used to search such design spaces, but their performance and parameter settings can depend on the problem under consideration \cite{r6,r7}. Moreover, robotic-hand structure selection can involve heterogeneous quantities, including individual kinematic performance, implementation-related costs, inter-finger interactions, and structural constraints. A formulation capable of representing these quantities within a common discrete optimization model is therefore desirable.
Quadratic unconstrained binary optimization (QUBO) provides a mathematical framework for representing combinatorial optimization problems using binary variables and quadratic interactions \cite{i8,k5}. Pairwise relationships between binary design choices can be represented directly as quadratic terms, while constraints can be incorporated through penalty functions. QUBO problems can be solved using conventional optimization approaches and can also be mapped onto Ising models for execution using quantum annealing (QA) hardware \cite{i7,i9}. Recent developments in quantum and quantum-inspired computing have consequently motivated the formulation of engineering optimization problems in hardware-compatible binary forms \cite{n1}. Nevertheless, the practical performance of QA relative to classical optimization remains dependent on the problem formulation and computational setting \cite{q1,q2}.

QUBO and QA have been investigated for combinatorial problems including graph partitioning, maximum clique, vehicle routing, scheduling, portfolio optimization, materials design, and structural design optimization \cite{q1,q2,q3}. Engineering-design studies have also formulated discrete design decisions as QUBO problems for solution using QA or hybrid quantum--classical approaches \cite{k1,k2,k3}. In robotics, quantum and quantum-inspired optimization has been investigated primarily for motion- and task-level problems such as scheduling \cite{q5} and inverse kinematics \cite{q4}. In particular, inverse kinematics has been formulated as a QUBO problem by discretizing candidate joint angles and solving the resulting binary selection problem using quantum annealing \cite{q4}. These applications mainly optimize motion, posture, or task decisions for a predefined robotic structure. In contrast, the use of QUBO to represent the combinatorial selection of the kinematic structure itself, including interactions and compatibility among multiple constituent structures, remains less explored.

This study presents a potential-dexterity-based QUBO formulation for combinatorial optimization of robotic-hand kinematic structures. Potential dexterity-related metrics are first obtained from kinematic analysis without assuming a specific target object or contact model. Candidate thumb and finger structures are represented by binary variables, and individual finger performance, thumb-to-finger reachable-workspace overlap, one-hot design selection, and pairwise structural compatibility are integrated into a single quadratic objective function. This formulation enables heterogeneous kinematic evaluation terms and structural constraints to be represented within a common binary optimization model. The resulting 27-variable problem is intentionally maintained at a scale for which the complete one-hot-valid design space can be enumerated, allowing a reference optimum to be established and the effects of structural compatibility and numerical discretization to be examined. The same QUBO formulation is subsequently evaluated using simulated annealing as a classical heuristic baseline and direct quantum annealing hardware execution. These evaluations are intended to validate the formulation and its solver compatibility rather than to demonstrate a computational advantage of quantum annealing.

%%%%%%%%%%%%%%%%%%%%%%%%%%%%%%%%%%%%%%%%%%%%%%%%%%%%%%%%%%%%
\section{Methods}
\label{sec:methods}

\subsection{Robotic-hand kinematic design space}
\label{sec:design_space}

The robotic-hand design space was constructed to represent discrete changes in kinematic structure and interactions among multiple finger structures. Four design factors were considered: (1) the phalanx length ratio of the thumb, (2) the MCP joint position of the middle finger, (3) the DoF configuration of the fingers excluding the thumb, and (4) the application of palm DoFs. Individual finger performance, thumb-to-finger interaction, and DoF-related cost were incorporated into the optimization objective. In addition, the palm DoF locations were defined between the middle and ring fingers and between the ring and little fingers, allowing structural dependencies between the ring- and little-finger designs to be represented.

\begin{figure}[htbp]
    \centering
    \includegraphics[width=0.85\textwidth]{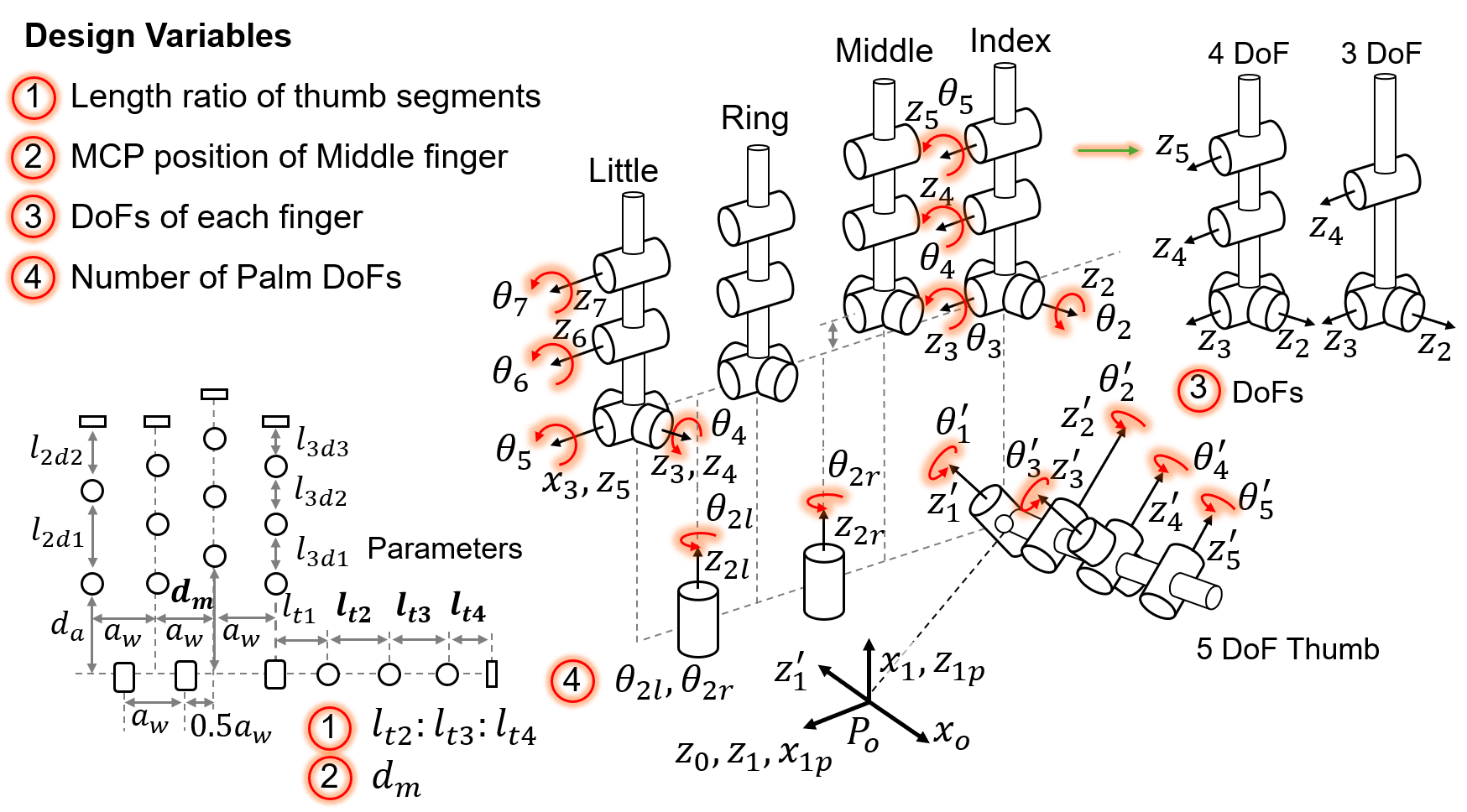}
    \caption{Overview of the robotic-hand kinematic structures, design variables, and kinematic parameters considered in the optimization problem.}
    \label{fig1}
\end{figure}

Figure~\ref{fig1} illustrates the robotic-hand kinematic structure, design variables, and kinematic parameters considered in this study. The design candidates of the thumb, index, middle, ring, and little fingers are denoted by $t$, $i$, $m$, $r$, and $l$, respectively.

The variable thumb segment lengths are denoted by $l_{t2}$, $l_{t3}$, and $l_{t4}$. The thumb DoF was fixed at five so that the thumb design variable was limited to the segment-length ratio. For $t_1$, the three variable segment lengths were assigned a ratio of 1:1:1. The remaining candidates $t_2$--$t_7$ were defined by ratios of 2:1:1, 1:2:1, 1:1:2, 2:2:1, 2:1:2, and 1:2:2, respectively.

For the MCP position, the reference configuration was defined such that all fingers were mounted at the same height. The mounting height along the $x_1$ direction with respect to the coordinate frame $P_o x_1 y_1 z_1$ was denoted by $d_a$. The positions of the other fingers were fixed, whereas the middle-finger position $d_w$ was varied. With the total hand length normalized to 1, three middle-finger positions were considered: $d_a$, $d_a+0.05$, and $d_a+0.10$.

Two DoF configurations were considered for each non-thumb finger. The four-DoF configuration consisted of one abduction/adduction (A/A) motion and three flexion/extension (F/E) motions, whereas the three-DoF configuration was obtained by removing one F/E motion. Accordingly, the index-finger candidates were defined as $i_1$ and $i_2$. For the middle finger, $m_1$ and $m_2$ correspond to the four- and three-DoF configurations at $d_w=d_a$, $m_3$ and $m_4$ correspond to those at $d_w=d_a+0.05$, and $m_5$ and $m_6$ correspond to those at $d_w=d_a+0.10$.

Two palm DoFs were considered. For the ring finger, $r_1$ and $r_2$ correspond to the four- and three-DoF finger configurations without the middle--ring palm DoF, whereas $r_3$ and $r_4$ correspond to the respective configurations with this palm DoF.

The little finger has eight candidates determined by the finger DoF configuration and the two palm DoF conditions. Candidates $l_1$, $l_2$, $l_5$, and $l_6$ are compatible with ring-finger designs without the middle--ring palm DoF, whereas $l_3$, $l_4$, $l_7$, and $l_8$ are compatible with ring-finger designs with the middle--ring palm DoF.

The total hand length was normalized to 1 for numerical simplification. According to the parameters illustrated in Fig.~\ref{fig1}, the index-finger dimensions satisfy

\begin{equation}
d_a+l_{3d1}+l_{3d2}+l_{3d3}=1.
\label{eq:hand_length}
\end{equation}

For the thumb, the sum of the three variable segment lengths was fixed at 0.60. The fixed offset $l_{t1}=0.10$ was included in the thumb kinematic model but excluded from the length-ratio design variable. Thus,

\begin{equation}
l_t=l_{t2}+l_{t3}+l_{t4}.
\label{eq:thumb_length}
\end{equation}

The distance from the index finger to the little finger was set to 0.54. The palm DoF was not intended to reproduce human metacarpal motion, but was modeled as a rotational DoF positioned at the center of the inter-finger spacing $a_w$. The resulting kinematic parameters are summarized in Table~\ref{tab1}.

\begin{table}[htbp]
\centering
\caption{Kinematic parameters of the robotic-hand model.}
\label{tab1}
\begin{tabular}{cccccc}
\hline
$d_a$ & $l_{3d1}$ & $l_{2d1}$ & $l_t$ & $l_{t1}$ & $a_w$ \\
\hline
0.46 & 0.18 & 0.27 & 0.60 & 0.10 & 0.18 \\
\hline
\end{tabular}
\end{table}

The design space was not intended to include all factors involved in practical robotic-hand design. Instead, representative kinematic structure-level variables were selected to construct a discrete design problem involving both individual finger characteristics and thumb-to-finger interactions.

\subsection{Kinematic analysis and potential-dexterity evaluation}
\label{sec:kinematic_analysis}

The design candidates were evaluated at the kinematic-structure level without assuming a specific task, target object, or contact model. Potential dexterity \cite{p4}, as adopted in the authors' previous studies \cite{p2,p3}, was used as the basis of the evaluation.

The evaluation factors considered in this study were global manipulability, DoF-based cost, and thumb-to-finger reachable-workspace overlap. The overlap workspace was evaluated using voxelized reachable regions of the thumb and each finger, whereas global manipulability and DoF-based cost were evaluated for individual finger designs.

The fingertip coordinates were represented with respect to the base frame $P_o x_o y_o z_o$ shown in Fig.~\ref{fig1}. The transformation matrix from link $i$ to link $i+1$ was defined as

\begin{equation}
T_i^{i+1}=
\begin{bmatrix}
R_i^{i+1} & p_i^{i+1}\\
0 & 1
\end{bmatrix},
\label{eq:transformation}
\end{equation}

\noindent where $R_i^{i+1}$ denotes the rotation matrix and $p_i^{i+1}$ denotes the displacement vector between adjacent links. Tables~\ref{tab2}--\ref{tab4} present the Denavit--Hartenberg (DH) parameters for the five-DoF thumb, the four-DoF middle finger, and the four-DoF little finger with two palm DoFs, respectively.

The thumb candidates were generated by modifying the corresponding segment lengths in Table~\ref{tab2}. The index, middle, ring, and little fingers without palm DoFs were generated by adjusting $a_w$ in Table~\ref{tab3}. For the middle finger, $d_a$ was additionally replaced by the corresponding $d_w$. Ring- and little-finger configurations affected by the palm DoFs were derived from Table~\ref{tab4} by including or removing the corresponding $\theta_{2r}$ and $\theta_{2l}$ terms and adjusting $a_w$ as required.

\begin{table}[htbp]
\centering
\caption{DH parameters of the five-DoF thumb.}
\label{tab2}
\begin{tabular}{ccccc}
\hline
$i$ & $\alpha_{i-1}$ & $a_{i-1}$ & $d_i$ & $\theta_i$ \\
\hline
1 & $-\pi/3$ & 0 & 0 & $\theta_1'$ \\
2 & $-\pi/2$ & $l_{t1}$ & 0 & $\theta_2'$ \\
3 & $\pi/2$ & $l_{t2}$ & 0 & $\theta_3'$ \\
4 & $-\pi/2$ & 0 & 0 & $\theta_4'$ \\
5 & 0 & $l_{t3}$ & 0 & $\theta_5'$ \\
6 & 0 & $l_{t4}$ & 0 & 0 \\
\hline
\end{tabular}
\end{table}

\begin{table}[htbp]
\centering
\caption{DH parameters of the four-DoF middle finger.}
\label{tab3}
\begin{tabular}{ccccc}
\hline
$i$ & $\alpha_{i-1}$ & $a_{i-1}$ & $d_i$ & $\theta_i$ \\
\hline
1 & 0 & 0 & $a_w$ & $\pi/2$ \\
2 & $\pi/2$ & $d_a$ & 0 & $\theta_2$ \\
3 & $-\pi/2$ & 0 & 0 & $\theta_3$ \\
4 & 0 & $l_{3d1}$ & 0 & $\theta_4$ \\
5 & 0 & $l_{3d2}$ & 0 & $\theta_5$ \\
6 & 0 & $l_{3d3}$ & 0 & 0 \\
\hline
\end{tabular}
\end{table}

\begin{table}[htbp]
\centering
\caption{DH parameters of the four-DoF little finger with two palm DoFs.}
\label{tab4}
\begin{tabular}{ccccc}
\hline
$i$ & $\alpha_{i-1}$ & $a_{i-1}$ & $d_i$ & $\theta_i$ \\
\hline
1 & $-\pi/2$ & 0 & 0 & $-\pi/2$ \\
2 & 0 & $1.5a_w$ & 0 & $\theta_{2r}$ \\
3 & 0 & $a_w$ & 0 & $\theta_{2l}$ \\
4 & 0 & $0.5a_w$ & $d_a$ & 0 \\
5 & $-\pi/2$ & 0 & 0 & $\theta_4-\pi/2$ \\
6 & $-\pi/2$ & 0 & 0 & $\theta_5$ \\
7 & 0 & $l_{3d1}$ & 0 & $\theta_6$ \\
8 & 0 & $l_{3d2}$ & 0 & $\theta_7$ \\
9 & 0 & $l_{3d3}$ & 0 & 0 \\
\hline
\end{tabular}
\end{table}

Using the DH convention, each finger was modeled as a serial kinematic chain \cite{c32}. Sequential multiplication of the transformation matrices gives the transformation from the base frame to the fingertip:

\begin{equation}
\begin{aligned}
p^e &= T_0^e p^o,\\
T_0^e &= T_0^1T_1^2\cdots T_{e-2}^{e-1}T_{e-1}^{e}.
\end{aligned}
\label{eq:forward_kinematics}
\end{equation}

The joint ranges used for the thumb and other fingers are summarized in Tables~\ref{tab5} and~\ref{tab6}. Because the number of finger DoFs varies among candidates, the joint variables were assigned according to motion type. The A/A motion corresponds to $\theta_4$ in Table~\ref{tab6}, the first F/E motion to $\theta_5$, and the remaining F/E motions to $\theta_6$ and $\theta_7$. These ranges were used to evaluate potential dexterity at the kinematic-structure level rather than to optimize the hand for a specific object or task.

\begin{table}[htbp]
\centering
\caption{Ranges of motion of the five-DoF thumb.}
\label{tab5}
\begin{tabular}{ccc}
\hline
$\theta_1'$ & $\theta_2',\theta_4',\theta_5'$ & $\theta_3'$ \\
\hline
$0$ to $\pi/2$ & $-\pi/2$ to $0$ & $-\pi/6$ to $\pi/6$ \\
\hline
\end{tabular}
\end{table}

\begin{table}[htbp]
\centering
\caption{Ranges of motion of the four-DoF finger with two palm DoFs.}
\label{tab6}
\begin{tabular}{ccccc}
\hline
$\theta_{2r}$ & $\theta_{2l}$ & $\theta_4$ & $\theta_5$ & $\theta_6,\theta_7$ \\
\hline
$0$ to $\pi/9$ &
$0$ to $\pi/6$ &
$-\pi/12$ to $\pi/12$ &
$-\pi/2$ to $\pi/9$ &
$-\pi/2$ to $0$ \\
\hline
\end{tabular}
\end{table}

The reachable workspace of each fingertip was generated by uniformly discretizing the corresponding joint ranges and computing the fingertip position at each sampled configuration through forward kinematics. The individual workspace volume was not directly used as an objective term. Instead, each reachable workspace was mapped onto a three-dimensional voxel grid, and overlap was defined when the reachable workspaces of the thumb and another finger occupied the same voxel. The resulting overlap workspace was used as a position-level measure of potential thumb--finger interaction. This metric represents positional co-reachability and does not account for fingertip orientation, approach direction, contact normal, link interference, or the existence of a simultaneous collision-free configuration.

Translational manipulability was evaluated using the manipulability measure \cite{c43}. The translational Jacobian $J(q)$ relates joint velocity to fingertip linear velocity according to

\begin{equation}
\dot{p}^{e}=J(q)\dot{q}.
\label{eq:jacobian}
\end{equation}

The manipulability at each sampled configuration was calculated as

\begin{equation}
w(q)=\sqrt{\det\left(J(q)J(q)^T\right)}.
\label{eq:manipulability}
\end{equation}

The global manipulability of each candidate was defined as the mean over the discretized joint space:

\begin{equation}
w_g=\frac{1}{N_q}\sum_{q\in\mathcal{Q}}w(q),
\label{eq:global_manipulability}
\end{equation}

\noindent where $\mathcal{Q}$ denotes the discretized joint space and $N_q$ denotes the number of sampled configurations. The joint-space discretization and voxelization-based numerical evaluation followed the methodology established in the authors' previous work \cite{p2}.

\subsection{Normalization of evaluation metrics}
\label{sec:normalization}

The manipulability, DoF-based cost, and overlap-workspace measures were normalized before construction of the QUBO coefficients.

Because the number of thumb DoFs was constant across all thumb candidates, only manipulability was considered in the individual thumb evaluation. The normalized thumb performance was defined as

\begin{equation}
S_T(t_i)=\frac{M_T(t_i)}{\max_i M_T(t_i)},
\label{eq:thumb_normalization}
\end{equation}

\noindent where $M_T(t_i)$ denotes the global manipulability of the $i$th thumb candidate.

For each non-thumb finger, manipulability was treated as a performance term and the number of DoFs as a cost-related term. The two terms were independently normalized and equally weighted:

\begin{equation}
\begin{aligned}
S_F(f_j)
&=0.5M_F^{\mathrm{norm}}(f_j)
-0.5D_F^{\mathrm{norm}}(f_j),\\
M_F^{\mathrm{norm}}(f_j)&=\frac{M_F(f_j)}{\max_jM_F(f_j)}, \quad D_F^{\mathrm{norm}}(f_j)=\frac{D(f_j)}{D_H},
\end{aligned}
\label{eq:finger_normalization}
\end{equation}

\noindent where $M_F(f_j)$ and $D(f_j)$ denote the global manipulability and DoF count of candidate $f_j$, respectively. The normalization constant $D_H=23$ represents the maximum number of independent DoFs of the complete hand among the considered design combinations.

Each palm DoF was counted only once in the DoF-based cost. The palm DoF between the middle and ring fingers was assigned to the ring-finger design, whereas the palm DoF between the ring and little fingers was assigned to the little-finger design. Therefore, ring-finger candidates containing the middle--ring palm DoF include one additional DoF in $D(f_j)$, and little-finger candidates containing the ring--little palm DoF include one additional DoF. An upstream palm DoF that also affects the kinematics and manipulability of a downstream finger was not counted again in the downstream finger's DoF cost.

The overlap-workspace values were normalized independently for the thumb--index, thumb--middle, thumb--ring, and thumb--little interaction blocks. For each finger type $f$, a single normalization denominator was defined using the maximum overlap among all combinations of the seven thumb candidates and all candidates of that finger:

\begin{equation}
O_{\mathrm{norm}}^{tf}(t_i,f_j)
=
\frac{O^{tf}(t_i,f_j)}
{\max_{i,j}O^{tf}(t_i,f_j)},
\label{eq:overlap_normalization}
\end{equation}

\noindent where $O^{tf}(t_i,f_j)$ denotes the overlap-workspace volume between thumb candidate $t_i$ and finger candidate $f_j$. Thus, all entries within a given thumb--finger interaction block share the same normalization denominator.

\subsection{QUBO formulation}
\label{sec:qubo}

The robotic-hand design problem was formulated as a combinatorial selection problem in which one candidate is selected for each finger group. The candidate sets are

\begin{equation}
\begin{aligned}
t &\in\{t_v\mid v=1,\ldots,7\}, &
i &\in\{i_w\mid w=1,2\},\\
m &\in\{m_x\mid x=1,\ldots,6\}, &r &\in\{r_y\mid y=1,\ldots,4\}, \quad l\in\{l_z\mid z=1,\ldots,8\}.
\end{aligned}
\label{eq:candidate_sets}
\end{equation}

Each candidate is represented by a binary decision variable,

\begin{equation}
t_v,i_w,m_x,r_y,l_z\in\{0,1\}.
\label{eq:binary_variables}
\end{equation}

The normalized individual evaluation coefficients for the thumb, index, middle, ring, and little fingers are denoted by $a_v$, $b_w$, $c_x$, $d_y$, and $e_z$, respectively. Because the QUBO objective is minimized, performance terms to be maximized were assigned negative coefficients. The individual-design reward was defined as

\begin{equation}
R_{\mathrm{des}}=-\sum_{v=1}^{7}a_vt_v-\sum_{w=1}^{2}b_wi_w-\sum_{x=1}^{6}c_xm_x-\sum_{y=1}^{4}d_yr_y-\sum_{z=1}^{8}e_zl_z.
\label{eq:rdes}
\end{equation}

Exactly one candidate must be selected from each design group. The corresponding one-hot penalty was

\begin{equation}
\begin{aligned}
P_{\mathrm{hot}}
=&\lambda_t\left(\sum_{v=1}^{7}t_v-1\right)^2
+\lambda_i\left(\sum_{w=1}^{2}i_w-1\right)^2\\
&+\lambda_m\left(\sum_{x=1}^{6}m_x-1\right)^2
+\lambda_r\left(\sum_{y=1}^{4}r_y-1\right)^2
+\lambda_l\left(\sum_{z=1}^{8}l_z-1\right)^2.
\end{aligned}
\label{eq:onehot_penalty}
\end{equation}

The normalized thumb-to-finger overlap-workspace interactions were represented by

\begin{equation}
R_{\mathrm{ovr}}=-\sum_{v=1}^{7}t_v\Bigg(\sum_{w=1}^{2}O_{vw}^{ti}i_w+\sum_{x=1}^{6}O_{vx}^{tm}m_x
+\sum_{y=1}^{4}O_{vy}^{tr}r_y+\sum_{z=1}^{8}O_{vz}^{tl}l_z\Bigg).
\label{eq:overlap_reward}
\end{equation}

The ring--little structural compatibility condition was incorporated as a quadratic penalty. For $r_1$ and $r_2$, the compatible little-finger candidates are $l_1$, $l_2$, $l_5$, and $l_6$, whereas for $r_3$ and $r_4$, the compatible candidates are $l_3$, $l_4$, $l_7$, and $l_8$. The incompatibility penalty was therefore defined as

\begin{equation}
P_{\mathrm{int}}=\lambda_{rl}\Big[(r_1+r_2)(l_3+l_4+l_7+l_8)+(r_3+r_4)(l_1+l_2+l_5+l_6)\Big].
\label{eq:interaction_penalty}
\end{equation}

The complete objective function was

\begin{equation}
f_{\mathrm{obj}}
=
R_{\mathrm{des}}
+
R_{\mathrm{ovr}}
+
P_{\mathrm{hot}}
+
P_{\mathrm{int}}.
\label{eq:objective}
\end{equation}

For binary variables, $v_p^2=v_p$. Expansion of the one-hot penalties therefore yields linear and quadratic terms together with an additive constant. The total constant generated by the five one-hot penalties is

\begin{equation}
C=
\lambda_t+\lambda_i+\lambda_m+\lambda_r+\lambda_l
=54.
\label{eq:constant}
\end{equation}

Because this constant does not affect the minimizing solution, it was omitted from the QUBO matrix. The QUBO energy used in all subsequent exhaustive, SA, and QA evaluations was therefore

\begin{equation}
E_{\mathrm{QUBO}}=f_{\mathrm{obj}}-C=\sum_{p=1}^{27}Q_{pp}v_p+\sum_{p=1}^{26}\sum_{q=p+1}^{27}Q_{pq}v_pv_q.
\label{eq:qubo_energy}
\end{equation}

The binary decision vector contains 27 variables and is ordered as

\begin{equation}
\mathbf{v}
=
[t_1,\ldots,t_7,
i_1,i_2,
m_1,\ldots,m_6,
r_1,\ldots,r_4,
l_1,\ldots,l_8].
\label{eq:decision_vector}
\end{equation}

An upper-triangular QUBO convention was used, such that each off-diagonal interaction coefficient appears once. Expansion of a one-hot penalty with coefficient $\lambda$ contributes $-\lambda$ to each corresponding diagonal coefficient and $2\lambda$ to each within-group off-diagonal pair. The remaining off-diagonal entries represent the thumb-to-finger overlap interactions and ring--little compatibility penalties. All unspecified entries were set to zero.

\subsection{Numerical and penalty settings}
\label{sec:numerical_settings}

For the baseline evaluation, all joint ranges were discretized at $5^\circ$ ($\pi/36$) intervals, and a voxel size of 0.05 was used for the overlap-workspace calculation.

Each one-hot penalty coefficient was set to twice the number of candidates in the corresponding design group. The ring--little interaction coefficient $\lambda_{rl}$ was set to the sum of the one-hot penalty coefficients for the ring- and little-finger groups. These coefficients were selected conservatively to discourage constraint violations relative to the normalized reward terms and were not optimized for QPU hardware. The resulting coefficients are summarized in Table~\ref{tab7}.

\begin{table}[htbp]
\centering
\caption{Penalty coefficients used in the QUBO formulation.}
\label{tab7}
\begin{tabular}{cccccc}
\hline
$\lambda_t$ &
$\lambda_i$ &
$\lambda_m$ &
$\lambda_r$ &
$\lambda_l$ &
$\lambda_{rl}$ \\
\hline
14.0 & 4.0 & 12.0 & 8.0 & 16.0 & 24.0 \\
\hline
\end{tabular}
\end{table}

\subsection{Exhaustive evaluation and discretization sensitivity}
\label{sec:exhaustive_method}

The complete one-hot-valid design space was exhaustively evaluated to establish a reference optimum for subsequent SA and QA evaluations and to examine the effect of the ring--little structural compatibility constraint. The five candidate groups contain 7, 2, 6, 4, and 8 candidates, respectively, yielding

\begin{equation}
7\times2\times6\times4\times8=2688
\end{equation}

one-hot-valid design combinations. Application of the ring--little compatibility condition reduces this set to 1344 compatible combinations.

The sensitivity of the design ranking to numerical discretization was evaluated by varying the joint-space sampling interval from $10^\circ$ to $2.5^\circ$ at a fixed voxel size of 0.05. The voxel size was separately varied from 0.10 to 0.025 at a fixed joint-space sampling interval of $5^\circ$. For each condition, the kinematic metrics were recalculated, the corresponding QUBO matrix was reconstructed, and the compatible design space was exhaustively evaluated.

\subsection{Simulated annealing}
\label{sec:sa_method}

Simulated annealing (SA) was used as a classical heuristic baseline for the QUBO problem constructed under the baseline discretization condition. The optimization was performed in Python using the \textit{SimulatedAnnealingSampler} \cite{p5}.

The number of reads (NoR), defined as the number of samples generated for the QUBO problem within each independent run, was varied among 500, 1000, 2000, and 5000. For each NoR condition, 50 independent runs were performed. The best solution obtained in each run was used to evaluate the best-found energy and the frequency with which individual low-energy solutions were recovered.

\subsection{Quantum annealing}
\label{sec:qa_method}

The same QUBO formulation was executed directly on a D-Wave quantum annealing processor through the D-Wave Leap cloud platform. The direct QPU solver \textit{Advantage\_system6} was used with the D-Wave Ocean SDK.

The QUBO was submitted using \textit{EmbeddingComposite(DWaveSampler())}, which performs minor embedding of the logical QUBO graph onto the physical QPU connectivity. Direct QPU sampling, rather than a hybrid quantum--classical solver, was used.

Except for NoR, the QPU and embedding parameters were not manually optimized. Minor embedding and chain strength were determined through the default Ocean workflow, and the default annealing time was used. These parameters were treated as execution conditions rather than tuned optimization parameters.

As in the SA evaluation, NoR values of 500, 1000, 2000, and 5000 were considered, with 50 independent QPU executions for each condition. The experiments were intended to evaluate direct hardware execution and recovery of low-energy feasible solutions from the formulated QUBO rather than to establish a computational advantage over classical optimization.

%%%%%%%%%%%%%%%%%%%%%%%%%%%%%%%%%%%%%%%%%%%%%%%%%%%%%%%%%%%%
\section{Results}
\label{sec:results}

\subsection{Exhaustive evaluation and discretization sensitivity}
\label{sec:exhaustive_results}

All energy values reported hereafter correspond to $E_{\mathrm{QUBO}}$ in Eq.~\eqref{eq:qubo_energy}, for which the constant offset $C=54$ is omitted.

\begin{figure}[htbp]
    \centering
    \includegraphics[width=0.8\textwidth]{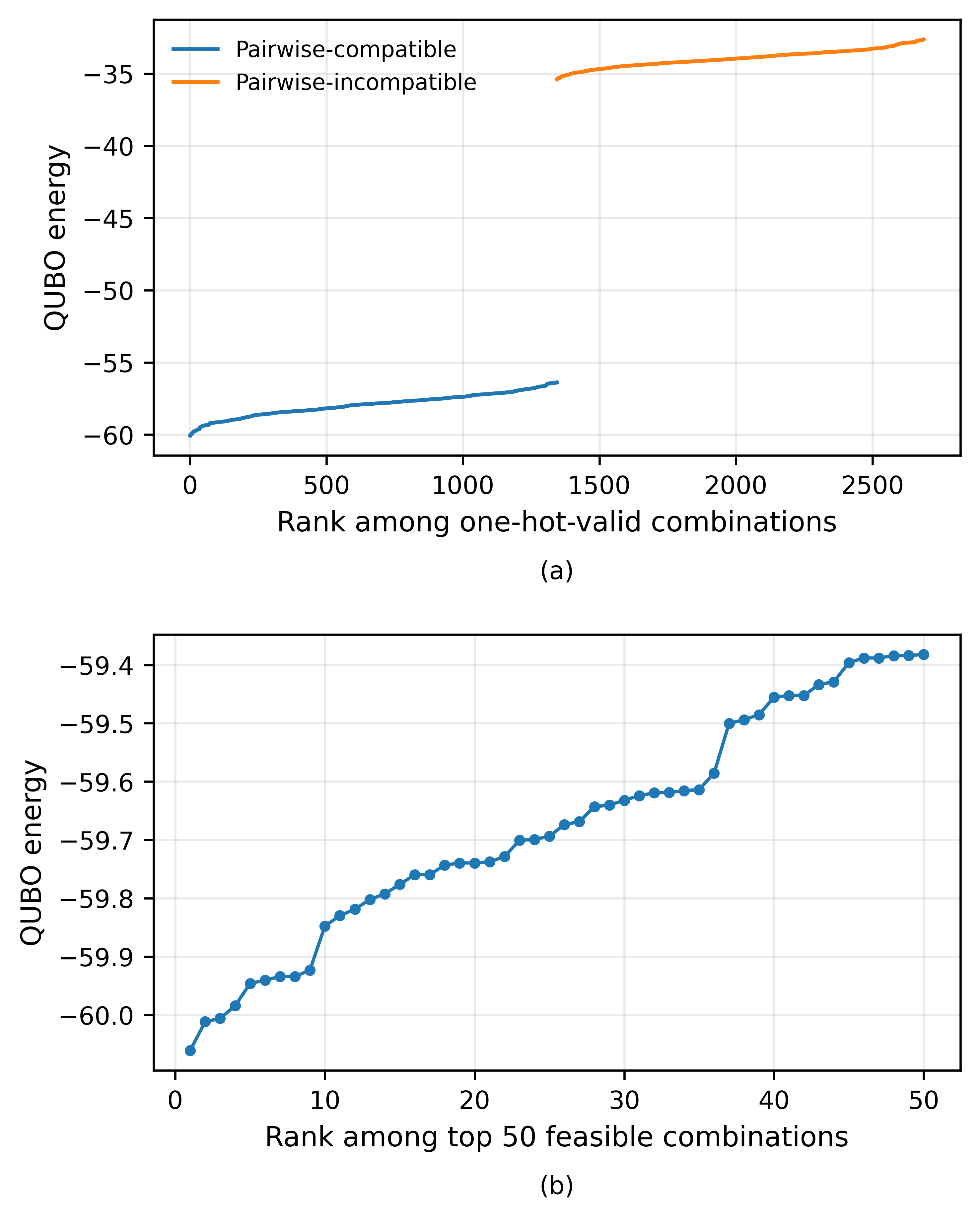}
    \caption{Exhaustive evaluation results. (a) Sorted QUBO energies of the 2688 one-hot-valid design combinations separated according to pairwise compatibility. (b) The 50 compatible design combinations with the lowest QUBO energies.}
    \label{fig6}
\end{figure}

Figure~\ref{fig6}(a) shows the QUBO energies of all 2688 one-hot-valid design combinations, separated according to whether the ring--little structural compatibility condition is satisfied. Pairwise-incompatible combinations receive the interaction penalty and are consequently shifted toward higher QUBO energies. Figure~\ref{fig6}(b) shows the 50 compatible design combinations with the lowest QUBO energies.

Under the baseline discretization condition of a $5^\circ$ joint-space sampling interval and a voxel size of 0.05, the minimum QUBO energy was $-60.061029$, corresponding to $(t_7,i_1,m_5,r_3,l_7)$. The second-lowest QUBO energy was $-60.011579$, corresponding to $(t_7,i_1,m_3,r_3,l_7)$. The energy difference between the two designs was therefore approximately 0.049451.

For the minimum-energy design $(t_7,i_1,m_5,r_3,l_7)$, the individual-design and overlap-workspace reward terms were $R_{\mathrm{des}}=-2.574501$ and $R_{\mathrm{ovr}}=-3.486529$, respectively. This design satisfies both the one-hot and structural compatibility constraints, such that $P_{\mathrm{hot}}=P_{\mathrm{int}}=0$. Consequently, $f_{\mathrm{obj}}=-6.061029$, corresponding to $E_{\mathrm{QUBO}}=-60.061029$ after omission of the constant offset.

Table~\ref{tab:sensitivity} summarizes the discretization sensitivity results. The baseline minimum-energy design $(t_7,i_1,m_5,r_3,l_7)$ was retained at joint-space sampling intervals of $10^\circ$, $5^\circ$, and $2.5^\circ$ with a voxel size of 0.05. It was also retained when the voxel size was reduced to 0.025 at a $5^\circ$ joint interval. At the coarser voxel size of 0.10, the middle-finger candidate changed from $m_5$ to $m_3$, whereas the other four selected candidates remained unchanged.

\begin{table}[htbp]
\centering
\caption{Discretization sensitivity of the exhaustive design ranking.}
\label{tab:sensitivity}
\begin{tabular}{ccc}
\hline
Voxel size & Joint interval & Minimum-energy design \\
\hline
0.05  & $10^\circ$  & $(t_7,i_1,m_5,r_3,l_7)$ \\
0.05  & $5^\circ$   & $(t_7,i_1,m_5,r_3,l_7)$ \\
0.05  & $2.5^\circ$ & $(t_7,i_1,m_5,r_3,l_7)$ \\
0.10  & $5^\circ$   & $(t_7,i_1,m_3,r_3,l_7)$ \\
0.025 & $5^\circ$   & $(t_7,i_1,m_5,r_3,l_7)$ \\
\hline
\end{tabular}
\end{table}

Thus, the selected design was unchanged across the tested joint-space sampling intervals and at the finer voxel resolution, but changed under coarse voxelization. The baseline minimum-energy design $(t_7,i_1,m_5,r_3,l_7)$ was used as the reference optimum for the subsequent SA and QA evaluations.

\subsection{Simulated annealing results}
\label{sec:sa_results}

\begin{figure}[htbp]
    \centering
    \includegraphics[width=0.75\textwidth]{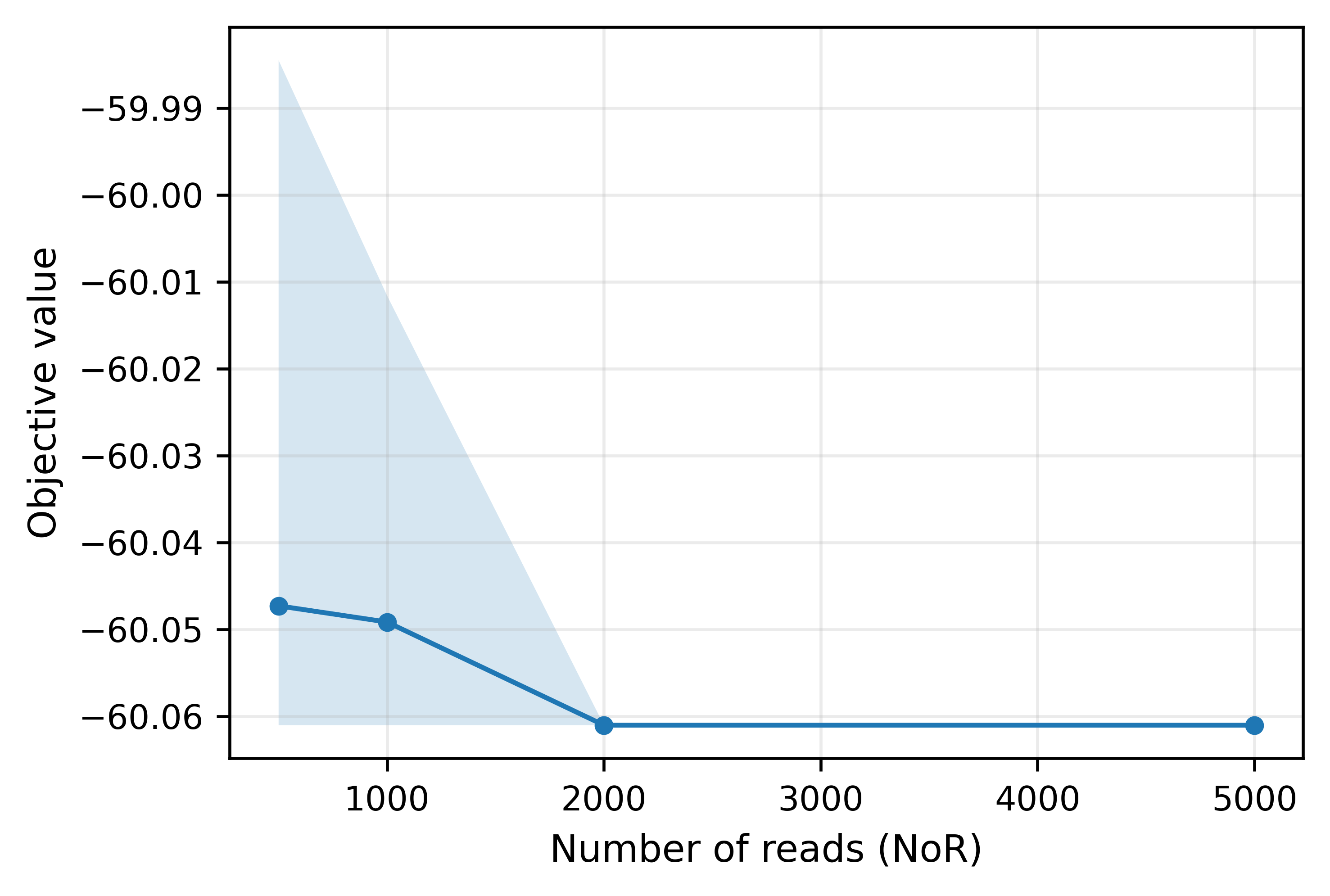}
    \caption{Best QUBO energies obtained using simulated annealing for different numbers of reads. The plotted values show the mean and minimum--maximum range across 50 independent runs.}
    \label{fig3}
\end{figure}

Table~\ref{tab8} summarizes the SA results for the four NoR conditions, and Fig.~\ref{fig3} shows the mean best QUBO energy and the corresponding minimum--maximum range across the 50 independent runs.

\begin{table}[htbp]
\centering
\caption{Simulated annealing results for different numbers of reads.}
\label{tab8}
\begin{tabular}{ccccc}
\hline
NoR & BFS$^1$ & BE$^2$ & MFOBS$^3$ & F$^4$ \\
\hline
500 &
$t_7,i_1,m_5,r_3,l_7$ &
-60.061029 &
$t_7,i_1,m_5,r_3,l_7$ &
37/50 \\
1000 &
$t_7,i_1,m_5,r_3,l_7$ &
-60.061029 &
$t_7,i_1,m_5,r_3,l_7$ &
38/50 \\
2000 &
$t_7,i_1,m_5,r_3,l_7$ &
-60.061029 &
$t_7,i_1,m_5,r_3,l_7$ &
50/50 \\
5000 &
$t_7,i_1,m_5,r_3,l_7$ &
-60.061029 &
$t_7,i_1,m_5,r_3,l_7$ &
50/50 \\
\hline
\multicolumn{5}{l}{\footnotesize BFS$^1$: Best-found solution; BE$^2$: Best energy.}\\
\multicolumn{5}{l}{\footnotesize MFOBS$^3$: Most frequently obtained best solution; F$^4$: Frequency of MFOBS.}
\end{tabular}
\end{table}

At NoR = 500, the best QUBO energies ranged from $-60.061029$ to $-59.984498$, with a mean of $-60.047301$, and the reference optimum $(t_7,i_1,m_5,r_3,l_7)$ was obtained as the best solution in 37/50 independent runs. Increasing NoR to 1000 produced only a small improvement in the mean best energy to $-60.049161$, while the reference-optimum frequency increased slightly to 38/50. A more distinct change was observed at NoR = 2000 and 5000, where all 50 runs reached the reference energy of $-60.061029$ and recovered the reference optimum. These results indicate that increasing NoR primarily reduced run-to-run variability and improved the reliability of optimum recovery, rather than changing the best attainable solution, which was already reached at NoR = 500. For the present QUBO instance, SA therefore exhibited stable recovery of the exhaustive reference optimum from NoR = 2000 onward.

\subsection{Quantum annealing results}
\label{sec:qa_results}

\begin{figure}[htbp]
    \centering
    \includegraphics[width=0.75\textwidth]{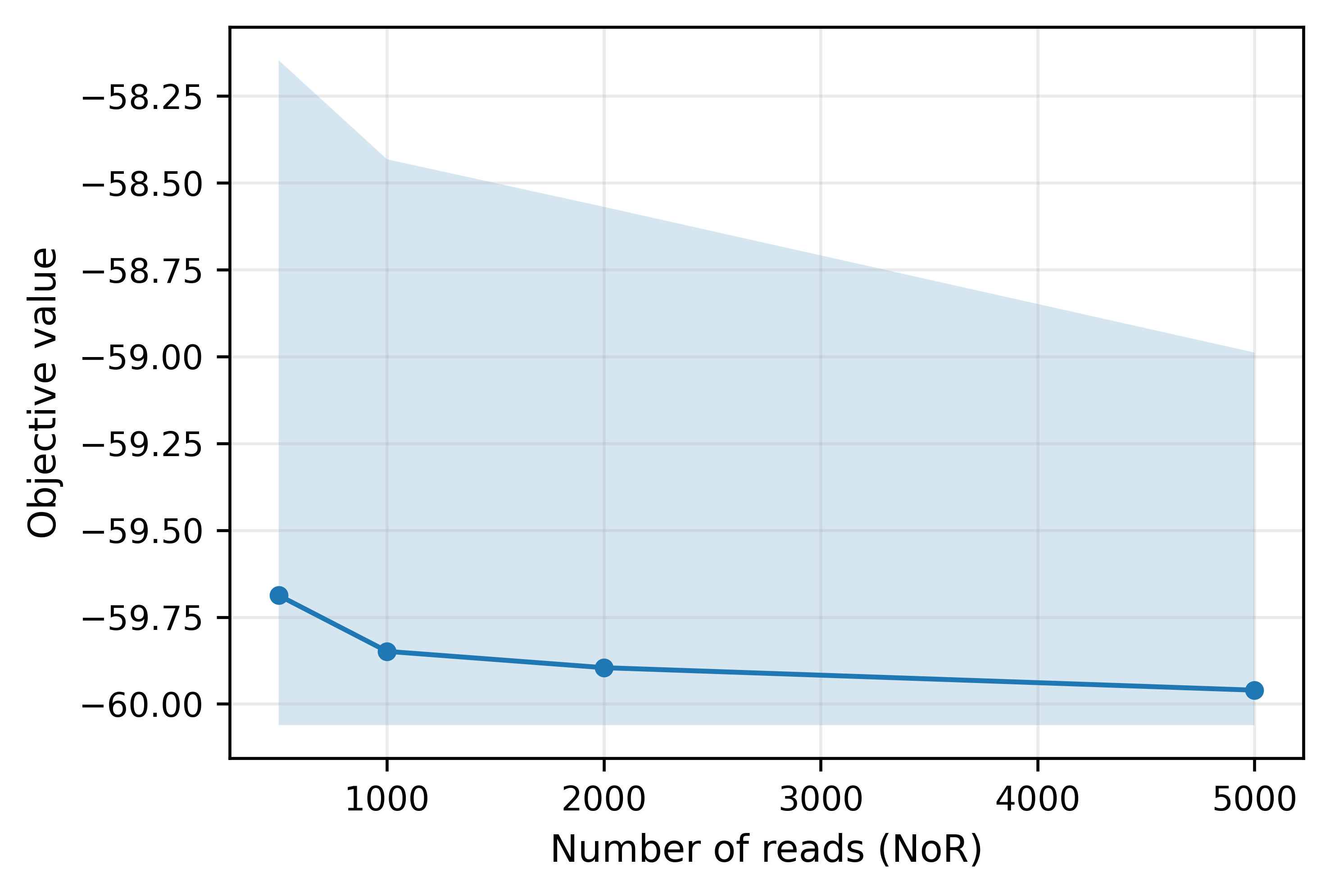}
    \caption{Best QUBO energies obtained using quantum annealing for different numbers of reads. The plotted values show the mean and minimum--maximum range across 50 independent QPU executions.}
    \label{fig4}
\end{figure}

All 50 QPU executions at each NoR condition yielded at least one feasible solution. The mean fractions of feasible samples among all returned reads were 0.2784, 0.2768, 0.2604, and 0.2534 for NoR = 500, 1000, 2000, and 5000, respectively. Table~\ref{tab9} summarizes the best-found and most frequently obtained best solutions, and Fig.~\ref{fig4} shows the mean best QUBO energies and their minimum--maximum ranges.

\begin{table}[htbp]
\centering
\caption{Quantum annealing results for different numbers of reads.}
\label{tab9}
\begin{tabular}{ccccc}
\hline
NoR & BFS$^1$ & BE$^2$ & MFOBS$^3$ & F$^4$ \\
\hline
500 &
$t_7,i_1,m_5,r_3,l_7$ &
-60.061029 &
$t_7,i_1,m_1,r_3,l_7$ &
5/50 \\
1000 &
$t_7,i_1,m_5,r_3,l_7$ &
-60.061029 &
$t_7,i_1,m_5,r_3,l_7$ &
7/50 \\
2000 &
$t_7,i_1,m_5,r_3,l_7$ &
-60.061029 &
$t_7,i_1,m_5,r_3,l_7$ &
14/50 \\
5000 &
$t_7,i_1,m_5,r_3,l_7$ &
-60.061029 &
$t_7,i_1,m_5,r_3,l_7$ &
21/50 \\
\hline
\multicolumn{5}{l}{\footnotesize BFS$^1$: Best-found solution; BE$^2$: Best energy.}\\
\multicolumn{5}{l}{\footnotesize MFOBS$^3$: Most frequently obtained best solution; F$^4$: Frequency of MFOBS.}
\end{tabular}
\end{table}

At NoR = 500, the best QUBO energies ranged from $-60.061029$ to $-58.146939$, with a mean of $-59.687444$, and the reference optimum $(t_7,i_1,m_5,r_3,l_7)$ was recovered as the best solution in 4/50 independent QPU executions. As NoR increased to 1000, 2000, and 5000, the mean best energy progressively approached the reference energy, reaching $-59.960392$ at NoR = 5000. Correspondingly, the reference-optimum recovery frequency increased from 4/50 to 7/50, 14/50, and 21/50. At NoR = 500, $(t_7,i_1,m_1,r_3,l_7)$ was the most frequently obtained best solution, whereas the reference optimum became the most frequent best solution from NoR = 1000 onward. These results indicate that increasing NoR improved the likelihood of obtaining lower-energy solutions and recovering the reference optimum, although run-to-run variability remained even at the highest tested NoR.

The mean fraction of feasible samples remained within approximately 25--28\% across the tested NoR conditions and did not increase with NoR. Thus, the improved recovery of the reference optimum with increasing NoR was associated with the larger number of samples available within each QPU execution rather than an increased proportion of feasible samples. The reference optimum was nevertheless observed under every tested NoR condition, demonstrating that the embedded QUBO could produce the exhaustive reference solution even at the lowest tested sampling count.

The hardware and embedding diagnostics are summarized in Table~\ref{tab10}. The 27-variable logical QUBO required approximately 74--75 physical qubits on average, with a maximum chain length of five. The mean chain strength was approximately 24.459, while the mean chain-break fraction remained low, ranging from 0.0175 to 0.0262 across the tested conditions. The annealing time was $20~\mu$s per sample.

\begin{table}[htbp]
\centering
\caption{Quantum annealing hardware and embedding diagnostics.}
\label{tab10}
\begin{tabular*}{\textwidth}{@{\extracolsep{\fill}}ccccccc}
\toprule
\multirow{2}{*}{NoR} &
Mean  &
Max.  &
Mean  &
Mean C$^2$- &
Annealing  &
Mean QPU  \\
 & PQ$^1$ & C$^2$ length & C$^2$ strength & break fraction & time ($\mu$s) & access time ($\mu$s)\\
\midrule
500  & 74.78 & 5 & 24.459 & 0.0262 & 20 & 70,084 \\
1000 & 74.92 & 5 & 24.459 & 0.0175 & 20 & 122,457 \\
2000 & 73.92 & 5 & 24.459 & 0.0191 & 20 & 230,188 \\
5000 & 74.48 & 5 & 24.459 & 0.0247 & 20 & 538,556 \\
\bottomrule
\multicolumn{7}{l}{PQ$^1$: Physical Qubits, C$^2$: Chain}
\end{tabular*}
\end{table}

The mean QPU access time increased from 70,084~$\mu$s at NoR = 500 to 538,556~$\mu$s at NoR = 5000 as more samples were requested per execution. Although the reference optimum was recovered under every tested NoR condition, the QA results remained distributed among several low-energy feasible solutions. The increased recovery frequency with increasing NoR therefore indicates an increased probability of observing the reference low-energy state as more samples were collected rather than deterministic convergence of the QPU to that state.

\subsection{Comparison of annealing-based optimization results}
\label{sec:solver_comparison}

\begin{figure}[htbp]
\centering
\includegraphics[width=0.85\textwidth]{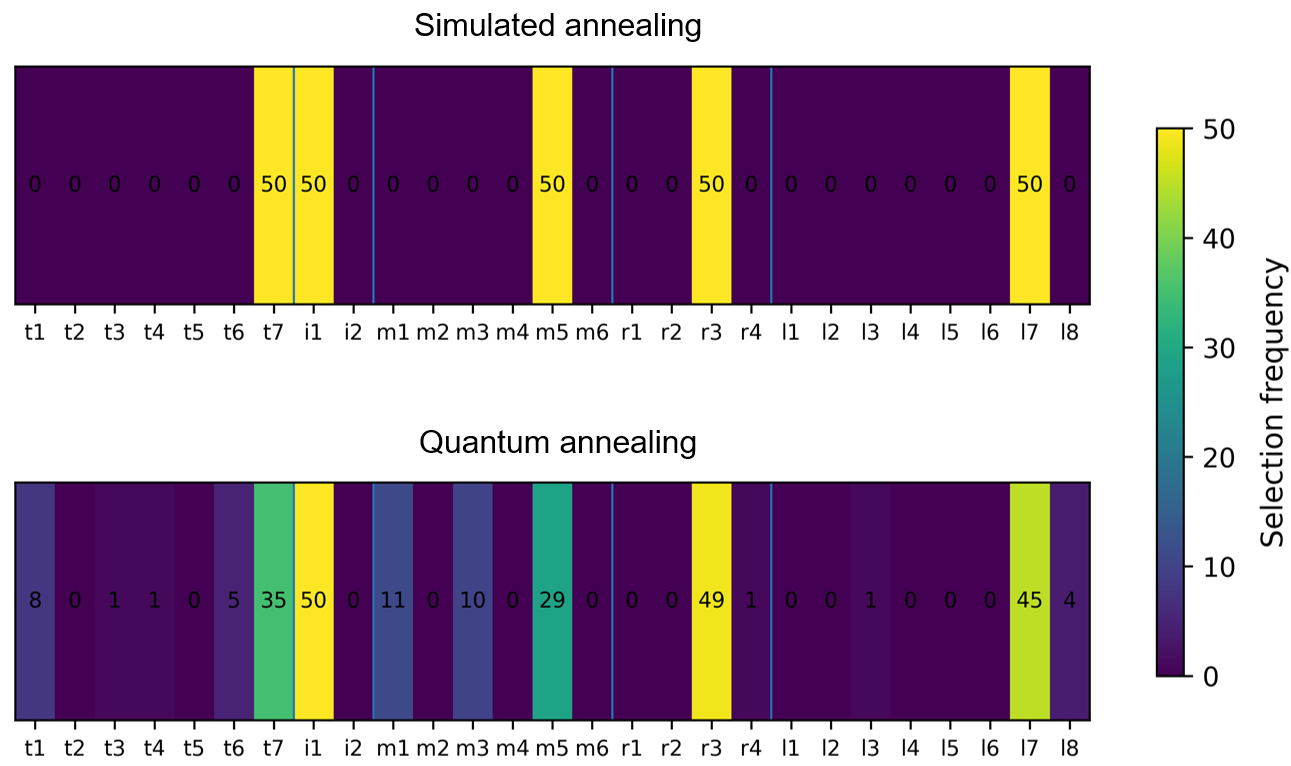}
\caption{Selection frequencies of individual design candidates obtained using simulated annealing and quantum annealing at NoR = 5000.}
\label{fig5}
\end{figure}

The SA and QA results were compared at NoR = 5000 using the exhaustive optimum as the reference. Figure~\ref{fig5} compares the selection frequencies of individual design candidates in the best solution obtained from each of the 50 independent runs or executions. These frequencies indicate the consistency of candidate selection across repeated runs and do not represent numerical distances between the categorical design indices.

For SA, all 50 independent runs recovered the reference optimum $(t_7,i_1,m_5,r_3,l_7)$. Consequently, $t_7$, $i_1$, $m_5$, $r_3$, and $l_7$ were each selected in all 50 runs. For QA, the complete reference-optimal combination was recovered in 21 of the 50 executions. At the individual-variable level, $t_7$, $i_1$, $m_5$, $r_3$, and $l_7$ were selected in 35, 50, 29, 49, and 45 executions, respectively. The remaining QA results were distributed mainly among alternative thumb- and middle-finger candidates, whereas the index-, ring-, and little-finger selections showed comparatively high consistency with the reference optimum.

Both annealing approaches therefore recovered the exhaustive reference optimum using the same QUBO formulation, although their run-to-run consistency differed under the tested conditions. SA recovered the reference optimum in all 50 runs at NoR = 5000, whereas QA recovered it in 21 of the 50 QPU executions. This comparison is used to evaluate recovery behavior for the common QUBO formulation and is not intended to establish a computational advantage of either solver.

%%%%%%%%%%%%%%%%%%%%%%%%%%%%%%%%%%%%%%%%%%%%%%%%%%%%%%%%%%%%
\section{Discussion}
\label{sec:discussion}

The results demonstrate that individual kinematic performance, DoF-related cost, thumb-to-finger workspace interactions, one-hot selection requirements, and pairwise structural compatibility can be represented within a common QUBO formulation for robotic-hand kinematic structure selection. Exhaustive enumeration provided a direct reference for interpreting the resulting QUBO energies and for evaluating the behavior of both annealing-based solvers.

The discretization sensitivity analysis provides an important qualification of the design result. The reference design was unchanged when the joint-space sampling interval was varied from $10^\circ$ to $2.5^\circ$ at a voxel size of 0.05 and was also unchanged when the voxel size was reduced from 0.05 to 0.025. In contrast, increasing the voxel size to 0.10 changed the selected middle-finger candidate from $m_5$ to $m_3$. The relatively small energy difference of approximately 0.0495 between the two leading designs under the baseline condition is consistent with the observed sensitivity of the middle-finger selection to coarse workspace voxelization. The selected structure should therefore be interpreted with respect to the numerical resolution used to evaluate the workspace interaction terms.

The SA and QA experiments served different validation purposes. SA provided a classical heuristic baseline for determining whether the low-energy state of the QUBO could be repeatedly recovered through stochastic optimization. At NoR values of 2000 and 5000, SA recovered the exhaustive reference optimum in every independent run. Direct QA execution demonstrated that the same logical QUBO could be embedded and sampled on physical quantum annealing hardware without reformulating the design problem. The reference optimum was observed under every tested NoR condition, but its recovery frequency remained lower than that obtained using SA. Because solver parameters and penalty coefficients were not optimized specifically for the QPU, these results should be interpreted as evidence of hardware-level executability and solution recovery rather than as a comparison of computational efficiency or quantum advantage.

The proposed optimization remains limited to kinematic-structure-level evaluation. In particular, the thumb-to-finger overlap metric represents positional overlap between voxelized fingertip reachable workspaces and does not establish simultaneous grasp feasibility. Fingertip orientation, approach direction, contact normal, collision-free simultaneous configurations, link interference, and contact or force conditions were not included. Practical robotic-hand design additionally requires consideration of finger thickness, actuator placement, required torque, transmission mechanisms, size and weight, manufacturing feasibility, and economic cost.

The present problem was deliberately restricted to a scale for which all 2688 one-hot-valid combinations could be enumerated, providing an exact reference within the defined discrete design space. For larger problems, scalability depends not only on the number of binary variables but also on QUBO connectivity. A one-hot group containing $n$ candidates can introduce up to $\binom{n}{2}$ pairwise interactions, and increased connectivity can increase minor-embedding complexity and physical-qubit requirements in direct quantum annealing. The present QPU results therefore do not establish large-scale scalability. Instead, they provide a validated starting point for extending the formulation to larger design spaces for which exhaustive enumeration is no longer practical.

%%%%%%%%%%%%%%%%%%%%%%%%%%%%%%%%%%%%%%%%%%%%%%%%%%%%%%%%%%%%
\section{Conclusion}
\label{sec:conclusion}

This study presented a potential-dexterity-based QUBO formulation for combinatorial optimization of robotic-hand kinematic structures. Potential dexterity-related metrics obtained from kinematic analysis were used to construct a discrete design-selection problem in which individual finger performance, DoF-related cost, position-level thumb-to-finger workspace overlap, one-hot selection constraints, and pairwise structural compatibility were represented within a unified binary quadratic objective function.

Exhaustive evaluation of the 2688 one-hot-valid combinations established $(t_7,i_1,m_5,r_3,l_7)$ as the reference optimum under the baseline discretization condition. The selected design remained unchanged across the tested joint-space sampling intervals and at a finer voxel resolution, whereas coarse voxelization changed the selected middle-finger candidate. SA consistently recovered the reference optimum at higher numbers of reads, and direct execution on D-Wave quantum annealing hardware also recovered the reference optimum under all tested NoR conditions.

The formulation therefore provides a common QUBO representation through which constrained robotic-hand kinematic structure selection can be evaluated using exhaustive enumeration, classical annealing, and direct quantum annealing. Future work will extend the design space to additional kinematic and mechanical variables and incorporate geometric, collision, force, and task-related constraints required for physically implementable robotic-hand designs.

\bibliographystyle{IEEEtran}
\bibliography{bibtex}

\end{document}